\documentclass[10pt,twocolumn,letterpaper]{article}

\usepackage[pagenumbers]{cvpr} %

\usepackage[dvipsnames]{xcolor}

\usepackage{times}
\usepackage{epsfig}
\usepackage{graphicx}
\usepackage{amsmath}
\usepackage{amssymb}

\usepackage{times}
\usepackage{epsfig}
\usepackage{graphicx}
\usepackage{amsmath}
\usepackage{amssymb}
\usepackage{siunitx}
\usepackage{multirow}
\usepackage{mycommands}
\usepackage{color}
\usepackage{pifont}

\usepackage{tabularx}
\usepackage{xpatch}

\usepackage{floatrow}
\newcommand\blfootnote[1]{%
  \begingroup
  \renewcommand\thefootnote{}\footnote{#1}%
  \addtocounter{footnote}{-1}%
  \endgroup
}
\usepackage[hang,flushmargin]{footmisc}

\definecolor{cvprblue}{rgb}{0.21,0.49,0.74}
\usepackage[pagebackref,breaklinks,colorlinks,citecolor=cvprblue]{hyperref}

\usepackage[capitalize]{cleveref}
\crefname{section}{Sec.}{Secs.}
\Crefname{section}{Section}{Sections}
\Crefname{table}{Table}{Tables}
\crefname{table}{Tab.}{Tabs.}

\def\paperID{80} %
\def\confName{3DV\xspace}
\def\confYear{2024\xspace}

\begin{document}

\title{LFM-3D: Learnable Feature Matching Across Wide Baselines Using 3D Signals}

\author{
Arjun Karpur$^\dagger$ \and
Guilherme Perrotta$^{\ast\dagger}$ \and
Ricardo Martin-Brualla$^\dagger$ \and
Howard Zhou$^\dagger$ \hspace{1cm}
Andr\'{e} Araujo$^\dagger$ \\ \\
$^\dagger$Google Research \\
{
\tt\small (arjunkarpur,rmbrualla,howardzhou,andrearaujo)@google.com}
\and
$^\ast$University of Campinas (UNICAMP)\\
{
\tt\small gperrotta@google.com}
}

\maketitle

\begin{abstract}
   
Finding localized correspondences across different images of the same object is crucial to understand its geometry.
In recent years, this problem has seen remarkable progress with the advent of deep learning-based local image features and learnable matchers. Still, learnable matchers often underperform when there exists only small regions of co-visibility between image pairs (\ie wide 
camera baselines).
To address this problem, we leverage recent progress in coarse single-view geometry estimation methods.
We propose \textbf{LFM-3D}, a Learnable Feature Matching framework that uses models based on graph neural networks and enhances their capabilities by integrating noisy, estimated 3D signals to boost correspondence estimation. 
When integrating 3D signals into the matcher model, we show that a suitable positional encoding 
is critical to effectively make use of the low-dimensional 3D information.
We experiment with two different 3D signals - normalized object coordinates and monocular depth estimates - and evaluate our method on large-scale (synthetic and real) datasets containing object-centric image pairs across wide baselines.
We observe strong feature matching improvements compared to 2D-only methods, with up to +6\% total recall and +28\% precision at fixed recall.
Additionally, we demonstrate that the resulting improved correspondences lead to much higher relative posing accuracy for in-the-wild image pairs - up to 8.6\% compared to the 2D-only approach.

\end{abstract}

\section{Introduction}
\label{sec:intro}

\begin{figure}[t]
\begin{center}
   \includegraphics[width=0.7\linewidth]{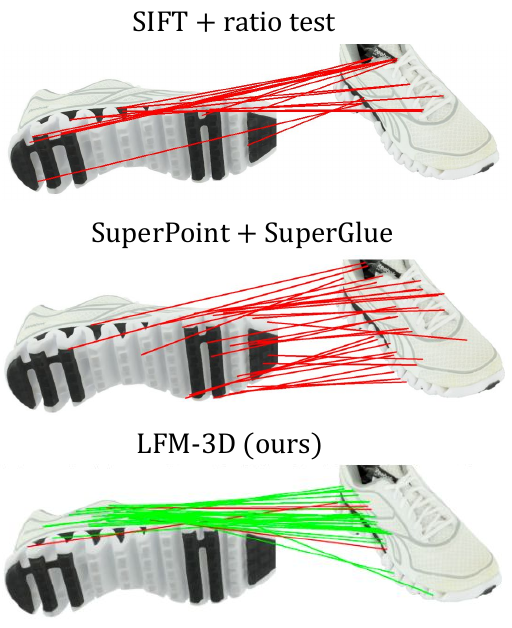}
\end{center}
   \caption{
   We propose \textbf{LFM-3D}, a novel learnable method for local feature matching leveraging 3D information.
   Infusing local feature matching with 3D signals enables accurate estimation of correspondences across very wide baselines, where conventional methods (SIFT + ratio test \cite{Lowe2004}) and even recent ones (SuperPoint + SuperGlue \cite{sarlin2020superglue}) fail -- we represent correct matches with green lines, and incorrect ones with red lines.
   Here, our method incorporates 3D normalized object coordinates as part of a graph neural network matcher, which significantly boosts the feature association process.
   }
\label{fig:key_fig}
\end{figure}

Matching images of the same object or scene is a core task in computer vision, since it enables the recovery of 3D structure and camera poses.
Many applications depend on these capabilities, ranging from augmented/virtual reality to novel view synthesis.
Central to image matching are local image features and associated algorithms.

\blfootnote{$^*$This work was completed while Guilherme was an intern at Google.}

Traditionally, hand-crafted techniques such as SIFT \cite{Lowe2004} or SURF \cite{bay2008speeded} were used to detect keypoints and describe them via local gradient statistics.
Putative feature matches were then obtained via simple nearest neighbor search, or its variants (\eg, Lowe's ratio test \cite{Lowe2004}), and fed into robust estimation methods such as RANSAC \cite{Fischler1981}.

In the past few years, deep learning based techniques have been proposed to replace hand-crafted methods in this area.
Several papers initially proposed to learn keypoint detectors \cite{moo2016learning,barroso2019key} or local descriptors \cite{mishchuk2017working,tian2017l2net,tian2019sosnet}.
Later, techniques were introduced to jointly model these two stages \cite{detone18superpoint,revaud2019r2d2,tyszkiewicz2020disk}, improving performance with end-to-end learning.
More recently, learnable feature matcher models were proposed to replace hand-crafted matching heuristics and obtain improved putative correspondences \cite{yi2018learning,sarlin2020superglue,sun2021loftr}.

Unfortunately, these methods still encounter difficulties in many cases. 
Wide baseline scenarios present a significant challenge due to small co-visibility regions and vastly different visual descriptors between the two keypoint sets.
Although learnable matchers integrate keypoint locations and global context to account for these factors, performance still suffers, as illustrated in~\figref{fig:key_fig}.
In this paper, we propose to address such challenging situations by integrating estimated 3D signals to help guide local feature matching even further. 
We propose using prior learned knowledge to infer rough estimates of the 3D structure of the underlying object. Then, along with the 2D keypoint locations, these 3D signals can help disambiguate matches when pixel-based descriptors are uninformative. Additionally, we propose using a positional encoding to upsample the 3D signal such that the matcher is more sensitive to the low-frequency signals.
Motivated by the increasing importance of object reconstruction, posing, and view synthesis~\cite{mildenhall2020nerf, truong2023sparf}, our experiments focus on improving object-centric image matching.

Depending on the availability of images and 3D models for given objects of interest, different types of 3D signals can be used.
In the case where a large-scale 3D model repository is available, we learn a class-specific Normalized Object Coordinate Space \cite{wang2019normalized} to predict 
3D coordinates for points on the object in a canonical orientation.
On the other hand, in the long-tailed case where no such data is available, we resort to class-agnostic monocular depth estimates \cite{Ranftl2020, ranftl2021vision} which can still provide helpful 3D hints for matching.
Thus, to evaluate our method, we perform an array of experiments, varying: 3D signals, object classes, controlled/in-the-wild datasets, and class-specific/generic variants of our model.
In more detail, our contributions are as follows.

\noindent\textbf{Contributions.}
\textbf{(1)} A novel method for local feature matching that can integrate different types of 3D information to help guide correspondence estimation, within a learnable framework that is a natural and simple extension to SuperGlue \cite{sarlin2020superglue}.
3D signals are associated with each local feature and input alongside descriptors and 2D positions into a graph neural network model that predicts correspondences. Single class variants of this method improves feature matching recall by more than 6\% and precision by up to 28\%. However, we find that our method can be scaled by jointly training on a mix of object classes, and that this generalized model can outperform both generic 2D matchers and class-specific LFM-3D.
Example results are highlighted in \figref{fig:key_fig}.
\textbf{(2)} We demonstrate that suitable positional encodings are required for this task in order to have 3D information actually help; simple encodings used in the learnable feature matching literature do not help much.
\textbf{(3)} As an example application of the improved correspondences, we present significant improvements in the downstream task of object relative pose estimation. In the large-scale Objectron dataset \cite{objectron2021} of in-the-wild images, relative pose estimates improve by up to $8.6$\% compared to the 2D-only approach.
We also outperform PnP-methods based on 3D coordinates by up to $10.4$\%. 

\section{Related Work}
\label{sec:rw}

\noindent\textbf{Traditional local feature extraction and matching.} Traditionally, hand-crafted methods, such as SIFT \cite{Lowe2004}, SURF \cite{bay2008speeded} or others \cite{tola2010daisy,chandrasekhar2009chog,calonder2012brief}, have been used to find correspondences between images.
Such keypoint detection approaches generally rely on finding salient low-level patterns such as blobs or corners, with representative methods being Difference-of-Gaussians \cite{Lowe2004}, Harris-Laplace or Hessian-Affine \cite{Mikolajczyk2004scale}.
Local description in this case is mainly based on gradient orientation statistics for small image patches around the keypoint.
In this conventional pipeline, associating local features across images generally builds on top of nearest neighbor matching, using Lowe's ratio test \cite{Lowe2004} or mutual consistency checks, and possibly additionally leveraging consensus methods \cite{tuytelaars2000wide,sattler2009scramsac}.

\noindent\textbf{Deep learning based local features.}
As the interest in deep learning methods for computer vision grew, hand-crafted feature extraction gave way to learned techniques.
As a first step, researchers proposed to independently learn either the keypoint detection \cite{verdie2015tilde,moo2016learning,mishkin2018repeatability,barroso2019key} or the keypoint description \cite{mishchuk2017working,luo2019contextdesc,tian2017l2net,tian2019sosnet,Simo2015discriminative,wang2020learning} stages.
This led to systems which were partially learned, but still relied on hand-crafted components.
A more recent trend is the unified modeling of detection and description, which enables end-to-end learning \cite{yi2016lift,noh2017large,detone18superpoint,ono2018lfnet,revaud2019r2d2,Dusmanu2019CVPR,tyszkiewicz2020disk}, leading to stronger results in image matching related tasks.
These methods may be trained using different levels of supervision: supervised at the feature level where the correspondence labels are computed from Structure-from-Motion pipelines \cite{yi2016lift,ono2018lfnet,revaud2019r2d2,Dusmanu2019CVPR,tyszkiewicz2020disk}, weaker supervision at the level of relative camera poses \cite{wang2020learning} or image-level labels \cite{noh2017large}, or finally fully self-supervised with synthetic image transformations \cite{detone18superpoint}.
Learned local features improve upon hand-crafted methods in several respects, \eg: the detection process extracts much more discriminative keypoints instead of hand-crafted low-level corners or blobs; the descriptor achieves more robust illumination and viewpoint invariance due to deep semantic representations.
Still, wide baseline matching and the presence of repeated patterns (\eg, building windows) may confuse these techniques and lead to poor correspondences.

\noindent\textbf{Learnable feature matching.}
One of the latest trends in the area of local image features is to additionally enable the learning of the feature matching process.
This goes beyond learned outlier filtering approaches \cite{yi2018learning,zhang2019learning}.
SuperGlue \cite{sarlin2020superglue} proposed graph neural network models in conjunction with optimal transport optimization in order to effectively associate sparse local features and reject outliers.
This method takes in sets of local features from two images, comprising local descriptors and 2D positions, and learns feature matching patterns with a data-driven approach.
A number of follow-up methods \cite{chen2021learning, shi2022clustergnn, lindenberger2023lightglue} improve matching accuracy and efficiency using updated GNN model architectures.
The learnable matching framework has also evolved to perform matching in a dense and semi-dense manner \cite{chen2022aspanformer, sun2021loftr, wang2022matchformer}, which avoids issues of potentially noisy keypoint detections at the cost of higher memory requirements.
In this work, we focus on the sparse case and extend the learnable feature matching model to consider 3D information at the keypoint level, allowing the framework to reason beyond 2D signals.

\noindent\textbf{Feature matching leveraging 3D information.}
Most local feature methods do not consider 3D information in the matching process.
Closest in spirit to our work is the method of Toft \etal \cite{toft2020single}, where monocular depth estimation is used to rectify perspective distortions, improving the quality of extracted local features.
In contrast, our work can leverage different types of 3D information in a learnable framework, which enables more precise matching. 
Another recent work in this area by Ma \etal \cite{ma2022virtual} relies on human poses as 3D priors in order to disambiguate occlusions during matching.
Our method shares some commonalities with it in the sense that both rely on some prior about the 3D shape of an object depicted in a 2D image.
However, we differ from them as our goal is to improve local feature representations for matching of visible parts, as opposed to their focus on associating invisible points for bundle adjustment.
Additionally, ObjectMatch~\cite{gumeli2023objectmatch} uses normalized object coordinates and local feature matches to register a collection of RGB-D frames against a target object. However, it relies on depth observations to align each camera and uses pairwise local feature matches as a peripheral signal during optimization over an alignment energy function. In contrast, our method operates directly on RGB-only image pairs and directly uses normalized coordinates to inform the feature matching model.
\section{LFM-3D}
\label{sec:method}

\subsection{Background}

We start by introducing the necessary background on learnable sparse feature matching, normalized object coordinates, and monocular 
depth estimation.

\noindent\textbf{Learnable sparse feature matching.}
Traditional matching methods rely solely on image region descriptors 
to propose correspondences. Learnable matchers improve upon these techniques 
by using additional information during the matching process, such as keypoint 
locations and global image context.
Our method builds on top of SuperGlue~\cite{sarlin2020superglue}, a landmark learnable sparse matching network, so we briefly introduce it here along with the required notation.

SuperGlue~\cite{sarlin2020superglue} is a graph neural network (GNN) model architecture that predicts correspondences between a pair of sparse keypoint sets.
Denote local features extracted from an image as $({\bf p},{\bf d})$, where $\mathbf p_i := (x, y, c)_i$ corresponds to the 2D position $x,y$ and confidence score $c$ for the $i$-th local feature, and $\mathbf d_i \in \mathbb R^D$ is the $i$-th local descriptor of dimension $D$.
A keypoint encoder combines $\mathbf p_i$ and $\mathbf d_i$ to produce a new matching representation $\mathbf x_i$ as follows:

\begin{equation}
    \label{eq:keypoint-encoder}
        \mathbf x_i = \mathbf d_i + \text{MLP}\left(\mathbf{p}_i\right)
\end{equation}

\noindent where MLP denotes a multi-layer perceptron that performs learnable positional encoding. Alternating layers of self-attention (within a 
single keypoint set) and cross-attention (between keypoint sets of two images) iteratively 
refine $\mathbf x_i$ by providing global context for both images to 
the local region.

Training uses ground truth correspondence supervision at the local feature level, 
and an optimal transport optimization problem is leveraged to solve for a partial assignment 
between the two feature sets. Ground truth correspondences between pairs of images are recovered using either known homographies or complete camera and depth information.

\noindent\textbf{Normalized object coordinates.}
The Normalized Object Coordinate Space (NOCS)~\cite{wang2019normalized} is a 3D space defined over the unit cube that can be used to represent the geometry of all instances of an object class in a canonical orientation.
Given an image of an unseen object instance, a NOCS map can be computed that predicts 3D normalized coordinates $\mathbf n_i \in \mathbb R^3$ per pixel with respect to the canonical reference frame.
Note that a prerequisite to exploit NOCS is to register all objects in the category in the same coordinate space.
From the NOCS estimates, one can solve for the 6D object pose of the instance with respect to the canonical orientation by using the 2D-3D correspondences between image pixels and NOCS points to solve the Perspective-n-Point (PnP) problem \cite{kneip2011novel,haralick1994review}. However, we observe in our experiments that fine-grained pose estimation accuracy is highly sensitive to the quality of the predicted NOCS map.

\begin{figure*}[t]
\begin{center}
   \includegraphics[width=0.7\linewidth]{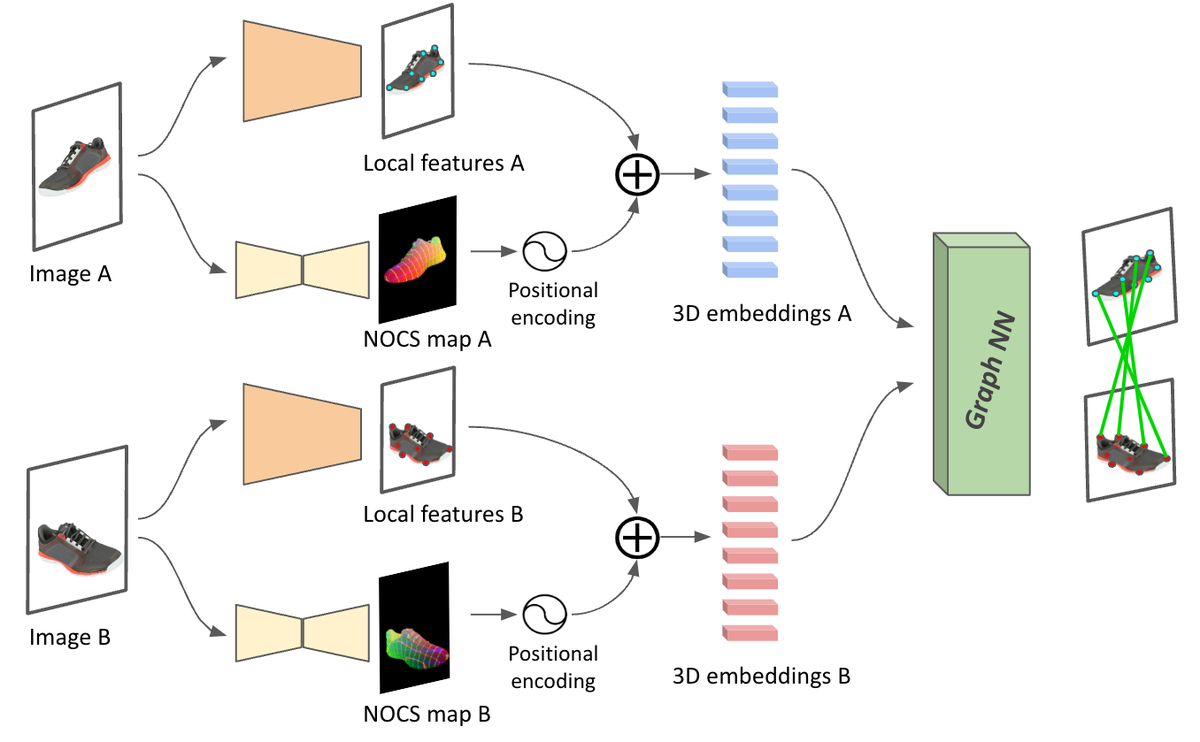}
\end{center}
   \caption{
\textbf{Block diagram of an instance of the proposed LFM-3D system, which uses NOCS~\cite{wang2019normalized} maps for 3D signals.}
We extract local features and normalized object coordinates (NOCS) from each image.
The NOCS maps are visualized by mapping XYZ to RGB.
The NOCS 3D coordinates undergo positional encoding and are combined with local features in order to generate 3D-infused local feature embeddings.
A graph neural network is then applied on these to propose correspondences.
Our method can find correspondences across images under very wide baselines, thanks to the 3D information leveraged from NOCS.
Besides NOCS, in this work we also instantiate the LFM-3D model with monocular depth estimates (MDE), which would follow the same process by changing NOCS maps to MDE maps.
   }
\label{fig:method}
\end{figure*}

\noindent\textbf{Monocular depth estimation.}
Monocular Depth Estimation (MDE)~\cite{eigen2014depth, godard2017unsupervised} is the task of predicting the depth value of each pixel given a single RGB image as the input. 
In most modern applications, the output is a dense inverse depth map of the same resolution as the input image. 
Compared to depth obtained from metric-based depth sensors (stereo cameras, LIDAR, etc.), MDE suffers from scale ambiguity, which makes it hard to be used directly for pose estimation. 
Despite the scale ambiguity problem, MDE models can be trained on massive amounts of data from varying image domains to produce robust relative depth estimates~\cite{Ranftl2020, ranftl2021vision}.
For our purpose, we found that these relative depth estimates provides sufficient 3D signals to boost correspondence estimation.

\subsection{Learning to match using 3D signals}

In the following, we present our method to integrate 3D signals in a learnable manner for improved feature matching, which is illustrated in \figref{fig:method}.

\noindent\textbf{Notation for 3D signals.}
Since many relevant 3D signals can be obtained as a dense feature map, to simplify the presentation, we assume that estimated information about the structure of the underlying object is extracted in this form.
Note, though, that sparse 3D signals could also naturally be incorporated into our framework.
Given an image $\textit{I} \in \mathbb R^{H \times W \times 3}$, the corresponding feature map $N \in \mathbb R^{H \times W \times V}, V \geq 1$ relates each image pixel to some vector $\mathbf{n}_i$ that encodes local information about the object's geometry. 
Given sufficient training data on the object class (i.e. 3D models), we can learn a strong prior over the object's geometry.
For example, a known object class with an accurate NOCS regression model available allows us to associate each pixel with an (estimated) 3D point on the object in the canonical orientation ($V = 3$). Under weaker assumptions, such as an unknown object class or limited 3D model data, monocular depth estimates may be used to relate each pixel to an estimated distance from the capturing camera ($V = 1$).

\noindent\textbf{Enhancing local features with 3D information.}
Given the feature map $N$, we use bilinear interpolation to look up the prediction for a local feature coordinate.
Thus, each local feature $i$  has information about its 2D pixel coordinates $\mathbf{p}_i$, local image content $\mathbf{d}_i$, and predicted 3D signal $\mathbf{n}_i \in \mathbb R^V$. 

\noindent\textbf{Extending SuperGlue to handle 3D signals.}
The 3D signals for each local feature can be directly used to improve each keypoint's matching representation:
\begin{equation}
    \label{eq:keypoint-encoder-3d}
        \mathbf x'_i = \mathbf d_i + \text{MLP}_{\text{2D}}\left(\mathbf{p}_i\right) + \text{MLP}_{\text{3D}}\left(\mathbf{n}_i\right)
\end{equation}

\noindent where $\text{MLP}_{\text{2D}}$ and $\text{MLP}_{\text{3D}}$ denote separate multi-layer 
perceptrons that act on the 2D keypoint information and the 3D signals, respectively. This 
3D-infused representation $\mathbf x'_i$ can be directly used in the GNN with the same architecture and loss functions to improve feature matching. Note that this design of summing up the features instead of concatenating allows for pre-training on datasets where no 3D information is available, as described below.

\noindent\textbf{Improved positional encoding.}
However, MLPs are know to have a spectral bias to learn overly smooth functions~\cite{tancik2020fourier}. This is particularly prevalent in models that learn functions over a distribution of low frequency, low dimensional values, such as from NOCS (3-dim) or inverse depth (1-dim). An alternative positional encoding formulation~\cite{tancik2020fourier, vaswani2017attention} can lift the input representation to a higher dimension space using periodic functions of increasing frequency. In our model, we use them to encode $\mathbf{n_i}$ before passing them onto the MLP. 
Our final representation for each local feature is then:

\begin{equation}
    \label{eq:keypoint-encoder-3d-pe}
        \hat{\mathbf x}_i = \mathbf d_i + \text{MLP}_{\text{2D}}\left(\mathbf{p}_i\right) + \text{MLP}_{\text{3D}}(\text{PE}\left(\mathbf{n}_i\right))
\end{equation}

\noindent where $\text{PE}$ stands for positional encoding. We use 10 frequencies with an implementation following the one of NeRF~\cite{mildenhall2020nerf}.
We refer to the new keypoint representations $\hat{\mathbf x}$ as \textit{3D-infused local 
feature embeddings}, or \textit{3D embeddings} for short.

\noindent\textbf{Multi-stage training.}
Limitations in object-centric data with the necessary match-level annotations~\cite{jampani2023navi} may cause the model 
to generalize poorly to different domains (e.g. from synthetic to real-world images). However, we 
note that there exists many multi-view and Structure-from-Motion datasets for outdoor scenes that 
can be used to initialize some layers of LFM-3D~\cite{li2018megadepth}. We bootstrap the LFM-3D model by training the 
$\text{MLP}_{\text{2D}}$ and $\text{GNN}$ layers of the matcher model from scratch for many 
iterations on synthetic homography and outdoor scene multi-view datasets.
Once the model has a strong understanding of the 2D 
correspondence estimation task, we introduce the $\text{MLP}_\text{3D}$ encoding layer and 
finetune all matcher-model layers jointly on object-centric data augmented with 3D estimates.
\section{Experiments}
\label{sec:experiments}
\subsection{Datasets}

We focus our experiments on object-centric image pairs taken from varying angles. We use synthetic renderings of 3D models for training, and evaluate on both synthetic and real-world image pairs. 

\noindent\textbf{Google Scanned Objects~\cite{downs2022google}.}
We use the high-quality scanned models with realistic textures provided by the Google Scanned Objects (GSO) dataset to train the learnable matcher components of LFM-3D. 
The GSO catalog features over 1,000 3D models of a diverse set of object categories, including \textit{shoes}, \textit{cameras}, \textit{toys}, \textit{bags}, \textit{car seats}, \textit{keyboards}, etc.
To define the canonical coordinate space for a category, we simply scale all the models so they have the same length (about $30cm$). Note that the 3D models were also prealigned during the capture process.

We render pairs of images with random poses, image dimensions, aspect ratios, focal lengths, and crops, and each rendered frame also provides ground truth camera matrices and depth maps.
To generate ground truth correspondence data, we extract local features from each image and use the camera/depth information to project a keypoint set from one image onto the other.
Ground truth matches are assigned via mutual nearest neighbors in reprojected pixel coordinate space, given a 3 pixel reprojection error threshold.

\noindent\textbf{Objectron~\cite{objectron2021}.}
The Objectron dataset contains object-centric videos with wide viewpoint 
coverage for nine object categories. Each video frame is 
labeled with camera intrinsics and extrinsics, and shows the instance in a 
real-world scene. To overcome the domain gap between synthetic and real-world 
images, we filter out all frames with $> 1$ object instance and use an off-the-shelf segmentation model to limit extracted features to the instance segmentation mask. The lack of metric depth data prevents us from training on Objectron frame pairs, but we leverage the freely available camera matrices for pose estimation evaluation purposes.

\noindent\textbf{Evaluation pairs.}
We sample image pairs at wide baselines to evaluate matcher performance in challenging scenarios. 
Specifically, we select two frames from the same 
scene and use the angular distance between the cameras' optical axes to 
filter for camera baselines between 90-120$^{\circ}$.

\subsection{Acquiring estimated 3D signals}

Given the flexible nature of LFM-3D, we provide experiments on two types of 3D estimates: NOCS maps and MDE inverse depth maps. The following section details how we retrieve these predicted 3D signals for each image pair.

\begin{figure}[t]
\begin{center}
 \begin{table}[H]
 \centering
 \vspace{-8mm}
 \resizebox{0.95\columnwidth}{!}{
    \begin{tabular}{cccccc}
    \parbox[t]{2mm}{\rotatebox[origin=c]{90}{\small  Input \quad\quad\quad}}&%
    \hspace{-2mm}\multirow{3}{*}{\includegraphics[width=0.18\linewidth]{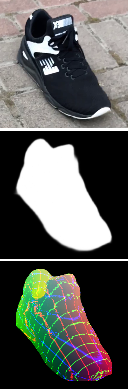}}&%
    \hspace{-4mm}\multirow{3}{*}{\includegraphics[width=0.18\linewidth]{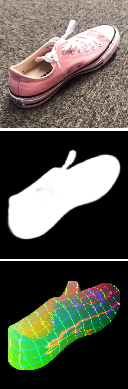}}&%
    \hspace{-4mm}\multirow{3}{*}{\includegraphics[width=0.18\linewidth]{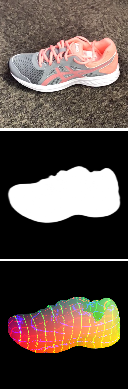}}&%
    \hspace{-4mm}\multirow{3}{*}{\includegraphics[width=0.18\linewidth]{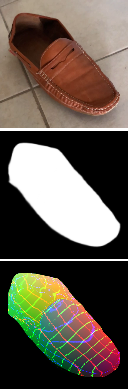}}&%
    \hspace{-4mm}\multirow{3}{*}{\includegraphics[width=0.18\linewidth]{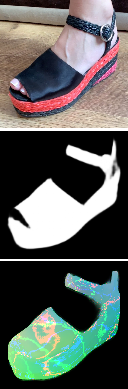}}
    \cr
    \parbox[t]{2mm}{\rotatebox[origin=c]{90}{\small Alpha \quad\quad\enskip}}\\
    \parbox[t]{2mm}{\rotatebox[origin=c]{90}{\small NOCS \quad\quad}}\\
    \end{tabular}
}
\end{table}%
\end{center}
\vspace{-25pt}
   \caption{
\textbf{Qualitative results of our trained NOCS model on Objectron.} Top: input image, middle: segmentation from off-the-shelf instance segmenter, bottom: NOCS rendering with axis aligned grid overlaid. Note that the NOCS model was trained only on synthetic renderings from Google Scanned Objects.
   }
\label{fig:nocs}
\end{figure}

\noindent\textbf{Normalized object coordinates.}
We opt to train a new NOCS map regression network for the \emph{shoe} object class, which contains enough 3D models for learning class-specific object coordinates.
As training data, we use the $327$ shoe models to produce 500 renderings each at resolutions of $256\times256$.
We randomly sample cameras while oversampling images with the object's up-vector aligned to the image's up-vector, and sample various crops, distances to the object and focal lengths, all while maintaining the object extent constant.
The network is a reduced-capacity version of Imagen's U-Net~\cite{saharia2022photorealistic} with fewer attention heads and features per layer, for a total of $28$M parameters.
Our model regresses the NOCS coordinates directly, in contrast to the classification approach~\cite{guler2018densepose} used in~\cite{wang2019normalized}.
Unlike~\cite{wang2019normalized}, we output NOCS coordinates at the same resolution as the input image.
We train the model with batch size $96$ for $300$k iterations on 8 NVIDIA Tesla V100 GPUs.

At inference time, given an image, we first apply an off-the-shelf instance segmenter for the target object category. Using the mask, we produce a tight crop around the object and pass it to the feed-forward network that estimates NOCS coordinates per pixel. 
On the synthetic dataset, the network is able to achieve a regression error of $1.4cm$ on a test set of unseen shoe models.
We show qualitative results of running NOCS on Objectron in Figure~\ref{fig:nocs}. Note that NOCS is only trained on synthetic renders from Google Scanned Objects, but is able to generalize to real images. We also note that the network fails for less common shoes (fifth column), as the dataset contains mostly sneaker type shoes.

\noindent\textbf{Monocular depth.}
We use an off-the-shelf, class-agnostic monocular depth estimation model to predict depth maps when studying long-tailed object categories where a large repository of 3D models is not available.
We use the publicly available DPT~\cite{ranftl2021vision} model, which is based on a Vision Transformer~\cite{dosovitskiy2020vit} backbone and trained on a large-scale meta-dataset across many image domains~\cite{Ranftl2020}. Following convention, we directly use the output inverse-depth maps to better handle depth values at large distances from the camera~\cite{civera2008inverse}.
Please see the appendix for qualitative depth estimate examples.

\subsection{Experimental setup}

\noindent\textbf{Evaluation scenarios and object classes.}
We focus our evaluation on three separate use cases of LFM-3D: a known object class with class-specific NOCS estimates, a known object class with class-agnostic MDE estimates, and unseen object classes with class-agnostic MDE estimates. 
In the `known' test object cases, we select the overlapping object classes between GSO and Objectron (\textit{shoe} and \textit{camera}) for evaluation, limiting training data to a single class in each case.
For the `unseen' test object evaluation, we use the entire GSO catalog to train generic (\ie, class-agnostic) model variants and evaluate on $5$ classes from Objectron: \textit{bike, book, cereal box, chair, laptop} (note that these classes do not overlap with those in the GSO catalog).

\noindent\textbf{Metrics.}
We report results on match-level statistics for the Google Scanned Objects dataset. We calculate precision/recall at the match level as the ratio of correct matches (precision) and ratio of ground truth matches recovered (recall) for a given correspondence prediction set.

Additionally, we evaluate correspondences on the downstream task of relative pose estimation, which is critical to many 3D reconstruction tasks. We estimate the essential matrix $E'$ that relates the two cameras using known camera intrinsics and predicted correspondences, and recover a relative rotation prediction. We calculate the relative rotation error between the predicted and ground truth rotation matrices using Rodrigues' formula and report accuracy within $5^{\circ}, 10^{\circ},$ and $15^{\circ}$ of error.

\begin{figure}[]
    \centering
            \vspace{-0.5em}

    \begin{subfigure}[b]{0.70\linewidth}
        \centering        

        \includegraphics[trim={1.2cm 0 2.5cm 1cm},clip,width=\textwidth]{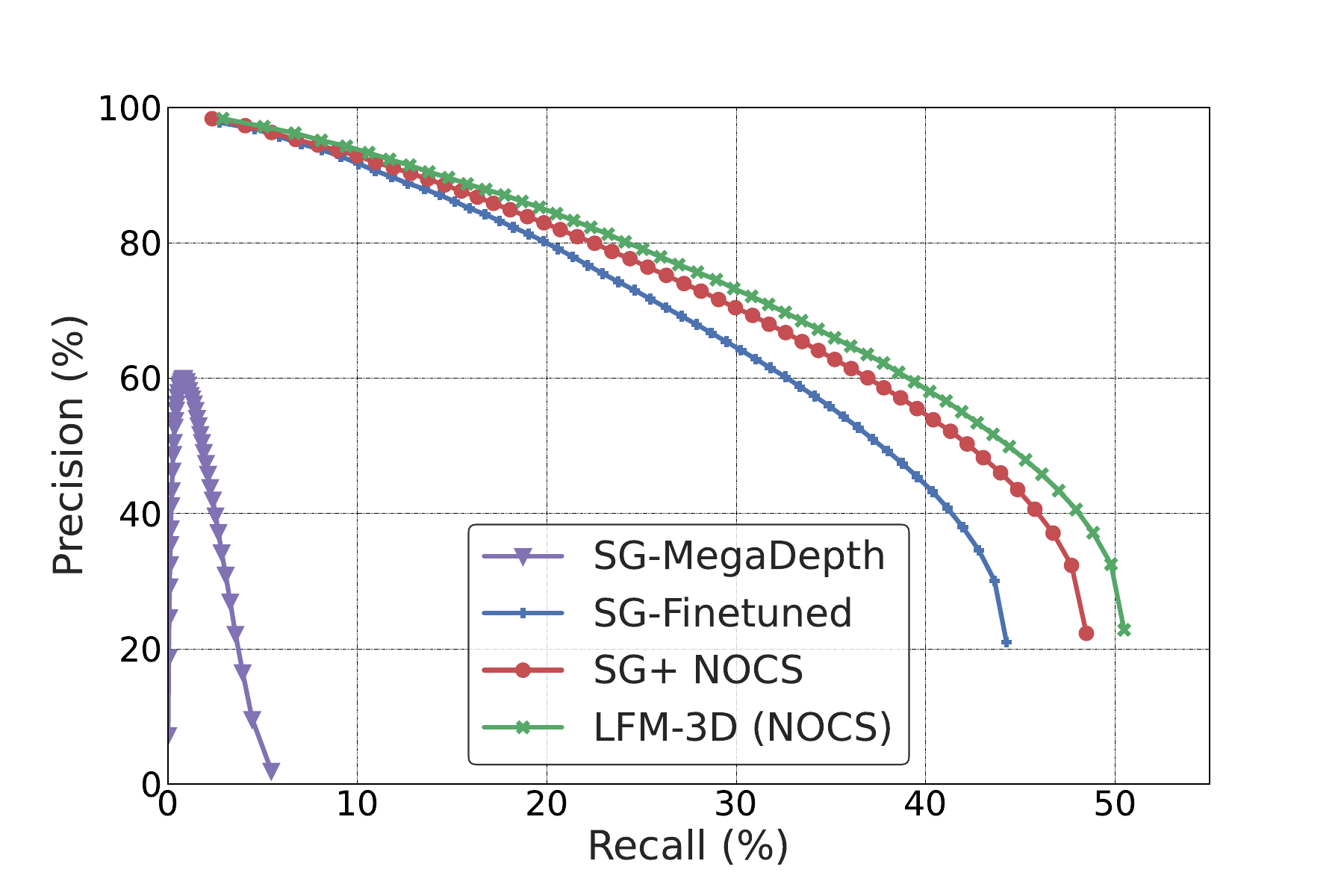}

        \caption{Shoe ($90^{\circ}$- $120^{\circ}$)}
        \label{fig:gso_pr_curves_a}
    \end{subfigure}

    \begin{subfigure}[b]{0.70\linewidth}
        \centering

        \includegraphics[trim={1.2cm 0 2.5cm 1cm},clip,width=\textwidth]{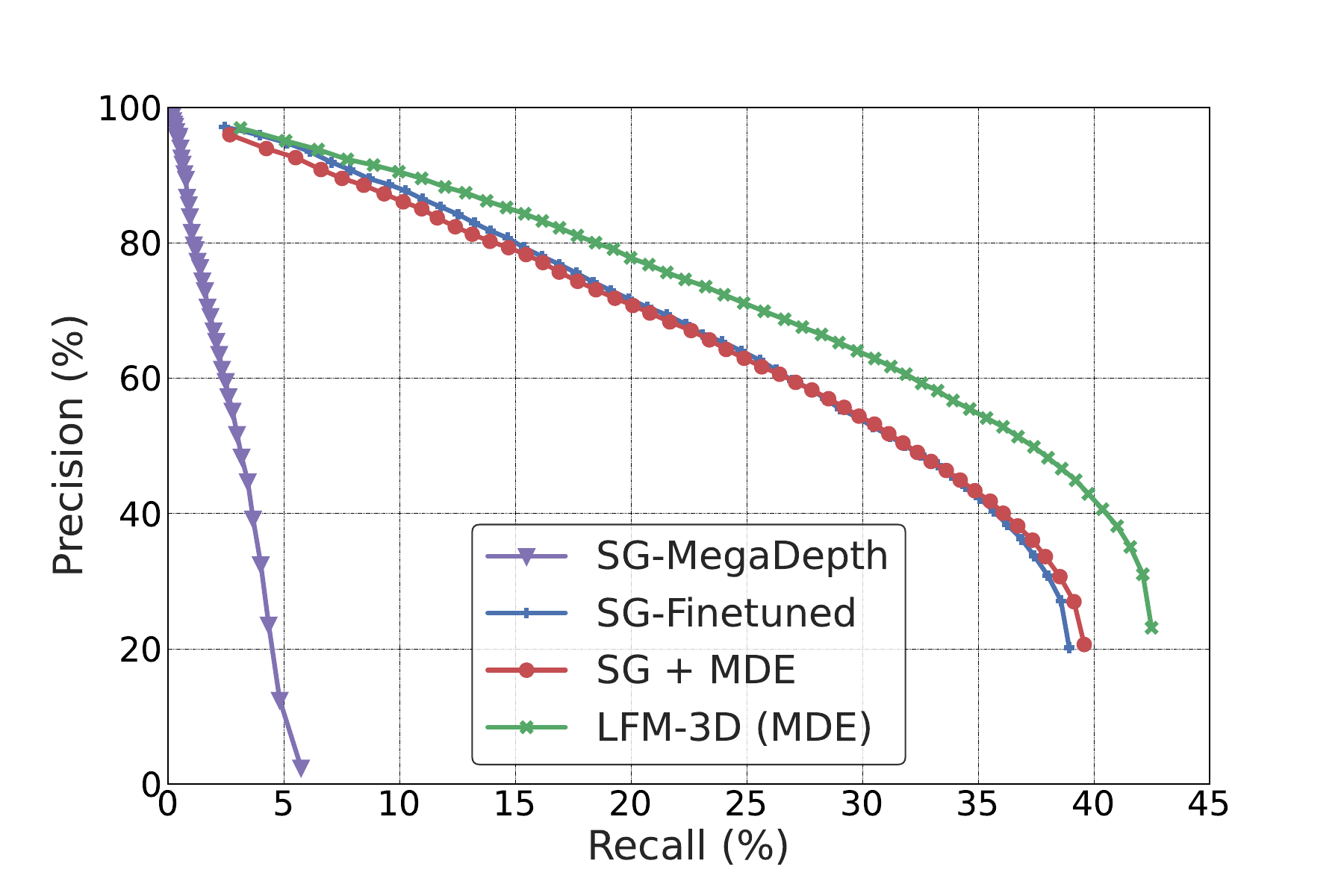}
        \caption{Camera ($90^{\circ}$- $120^{\circ}$)}
        \vspace{-0.5em}
        \label{fig:gso_pr_curves_b}

    \end{subfigure}
    \caption{\textbf{Correspondence-level precision/recall curves} for ablated version of class-specific LFM-3D on the Google Scanned Objects evaluation datasets.}
    \label{fig:gso_pr_curves}
\end{figure}

\begin{table*}
    \small
    \centering
    \resizebox{0.90\textwidth}{!}{
    \begin{tabular}{clccc|ccc|ccc}
        \toprule
        & & \multicolumn{3}{c}{\emph{Shoe}} & \multicolumn{3}{c}{\emph{Camera}} & \multicolumn{3}{c}{\emph{Unseen}} \\
        Row & Method & Acc@$5^{\circ}$ & Acc@$10^{\circ}$ & Acc@$15^{\circ}$ & Acc@$5^{\circ}$ & Acc@$10^{\circ}$ & Acc@$15^{\circ}$ & Acc@$5^{\circ}$ & Acc@$10^{\circ}$ & Acc@$15^{\circ}$\\
        \midrule
        
        1 & SIFT + MNN & 4.0 & 7.8 & 11.8 & 4.6 & 9.0 & 13.2 & 4.8 & 9.2 & 12.9 \\
        2 & SIFT + NN/Ratio & 4.4 & 8.7 & 13.2 & 4.3 & 9.0 & 13.0 & 4.4 & 8.5 & 12.5 \\
        \midrule
        
        3 & NOCS + Sparse-PnP & 4.8 & 18.4 & 33.4 & - & - & - & - & - & - \\
        4 & NOCS + Dense-PnP & 4.2 & 18.9 & 34.9 & - & - & - & - & - & - \\
        \midrule
        
        5 & LoFTR & 1.9 & 4.0 & 5.7 & 4.9 & 8.4 & 10.4 & 4.4 & 6.3 & 7.7 \\
        \midrule
        
        6 & SuperPoint + MNN & 3.4 & 7.1 & 11.1 & 2.7 & 5.9 & 8.8 & 5.9 & 10.7 & 14.8 \\
        7 & SuperPoint + NN/Ratio & 4.0 & 7.8 & 11.9 & 3.0 & 6.6 & 10.2 & 4.9 & 9.5 & 13.9 \\
        8 & SuperPoint + SG-MegaDepth & 2.6 & 5.6 & 8.6 & 2.7 & 5.5 & 8.6 & 6.4 & 9.8 & 12.5 \\
        9 & SuperPoint + SG-Finetuned \textit{(class specific)} & 11.0 & 22.2 & 31.6 & 10.6 & 21.4 & 29.2 & - & - & - \\
        10 & SuperPoint + SG-Finetuned \textit{(generic)} & 5.5 & 12.7 & 19.2 & 10.4 & 19.9 & 27.3 & 17.7 & 26.9 & 32.4 \\
        \midrule
        \midrule
        
        11 & SuperPoint + LFM-3D \textit{(class specific, NOCS)} & \textbf{14.3} & \textbf{29.3} & \textbf{40.2} & - & - & - & - & - & - \\
        12 & SuperPoint + LFM-3D \textit{(class specific, MDE)} & 7.4 & 16.3 & 24.2 & 11.8 & 22.4 & 30.7 & - & - & - \\
        \midrule
        13 & SuperPoint + LFM-3D \textit{(generic, MDE)}  & 6.3 & 15.1 & 22.8 & \textbf{13.0} & \textbf{25.7} & \textbf{34.3} & \textbf{19.1} & \textbf{28.9} & \textbf{34.5} \\
        \bottomrule
    \end{tabular}
    }
    
    \caption{\textbf{Relative pose estimation results on Objectron sampled video frames.} We calculate the error at multiple accuracy thresholds between the ground truth rotation matrix and recovered relative rotation matrix from essential matrix estimation.}
    \label{tab:rel_pose_results}
\end{table*}

\noindent\textbf{Compared techniques.}
We compare LFM-3D to popular 2D-based sparse matching techniques, such as SIFT~\cite{Lowe2004} features with heuristics-based matchers and SuperPoint~\cite{detone18superpoint} features with SuperGlue correspondence prediction.
For \emph{shoe} relative pose estimation, we can also compare against sparse matching between 2D keypoints and their 3D NOCS estimates (\ie Sparse PnP), and with dense matching between all 2D pixel coordinates and their 3D NOCS estimates (\ie Dense PnP).
Additionally, we compare our method to the popular dense learnable matching technique, LoFTR~\cite{sun2021loftr}, for relative pose estimation.

\noindent\textbf{Architecture and training details.}
Following \cite{sarlin2020superglue}, we use SuperPoint local features \cite{detone18superpoint} in our learnable matching framework.
We initialize the 2D keypoint encoder, graph neural network, and optimal transport layers from a 2D-only matching model pretrained on synthetic homographies and Megadepth~\cite{li2018megadepth} image pairs, following the recipe described in \cite{sarlin2020superglue}.
The model extracting 3D signals (\ie, NOCS or MDE) is frozen when training the LFM-3D matching layers.
We mirror the $\text{MLP}_{3D}$ architecture after $\text{MLP}_{2D}$, but increase the number of parameters in each layer as follows: $[64, 64, 128, 128, D]$, where $D$ is the local feature descriptor ($\mathbf d_i$) dimension.
An initial learning rate of \num{8e-5} is used until 65k iterations, after which the learning rate decays by a factor of $0.999995$ per iteration. The model was 
trained for 750,000 iterations with a batch size of 56 on the same set of GPUs as stated above. LFM-3D is implemented natively in Tensorflow 2~\cite{tensorflow2015-whitepaper} from scratch, and we use a public Tensorflow reproduction of Superpoint.

\begin{figure*}[t]

\begin{center}
    \begin{subfigure}[t]{0.78\textwidth}
        \hspace{-5pt}
        \includegraphics[width=\linewidth]{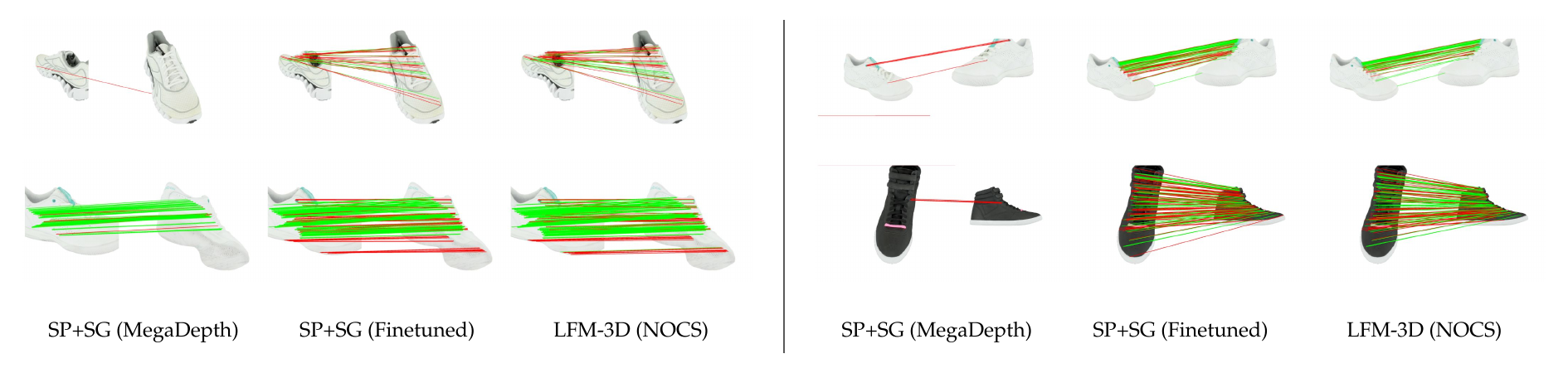}
        \caption{Google Scanned Objects wide-baseline pairs.\vspace{5pt}}
        \label{fig:qual_gso}
    \end{subfigure}
    
    \begin{subfigure}[t]{0.78\textwidth}
        \hspace{-5pt}
        \includegraphics[width=\linewidth]{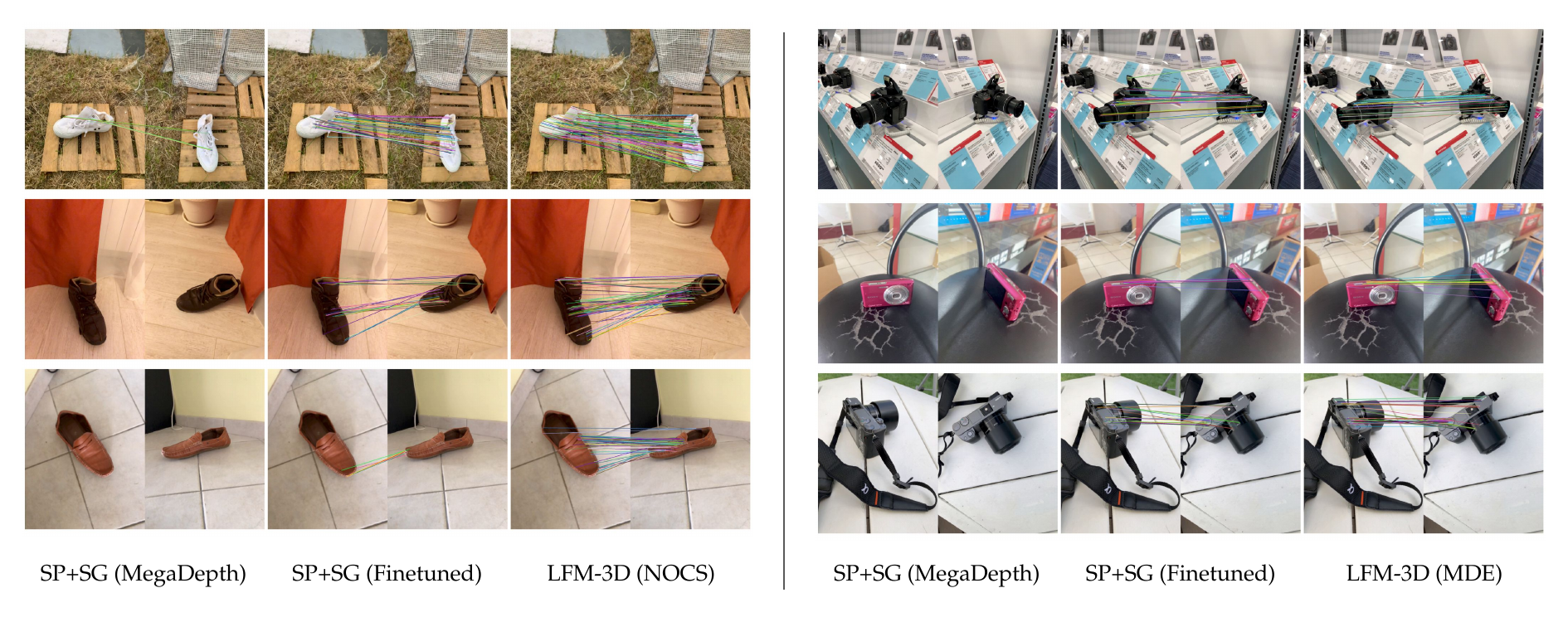}
        \caption{ Objectron wide-baseline image pairs. }
        \label{fig:qual_objectron}
    \end{subfigure}
    
    \begin{subfigure}[t]{0.45\textwidth}
        \vspace{7pt}
        \begin{tabular}{ccc}
            \includegraphics[width=0.32\linewidth]{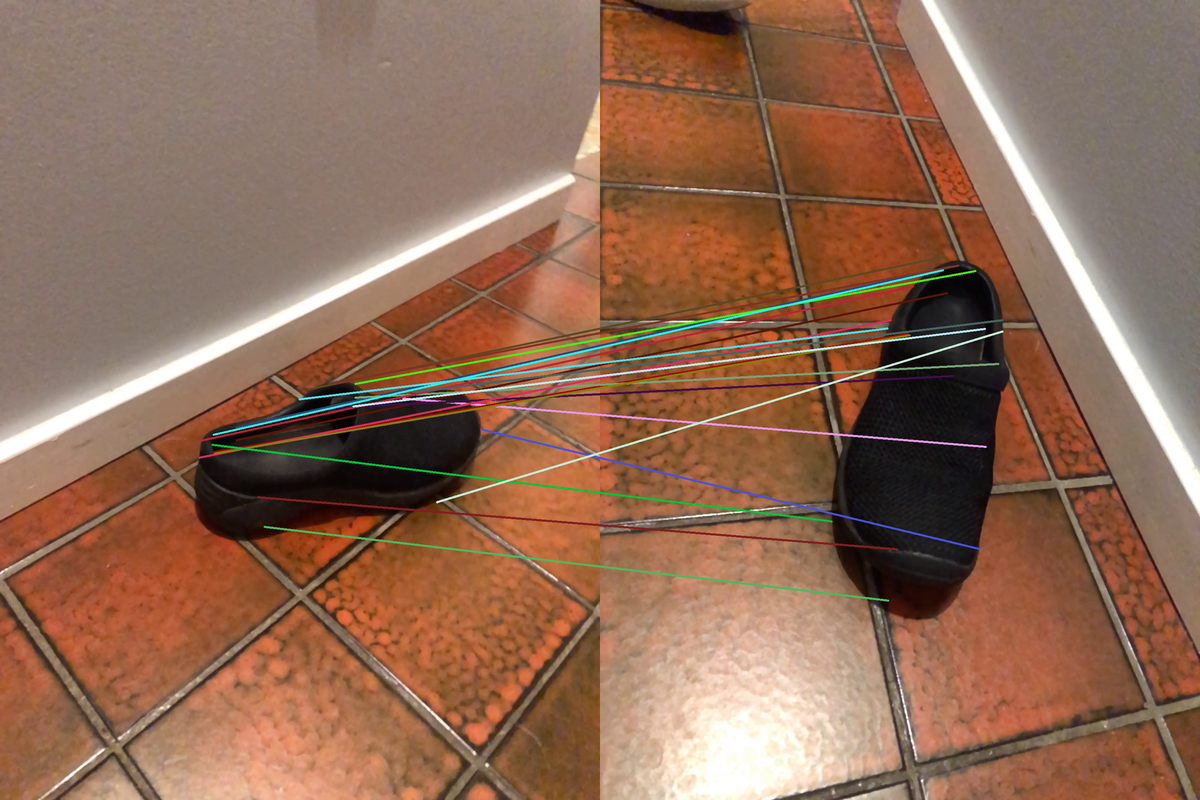} & 
            \includegraphics[width=0.32\linewidth]{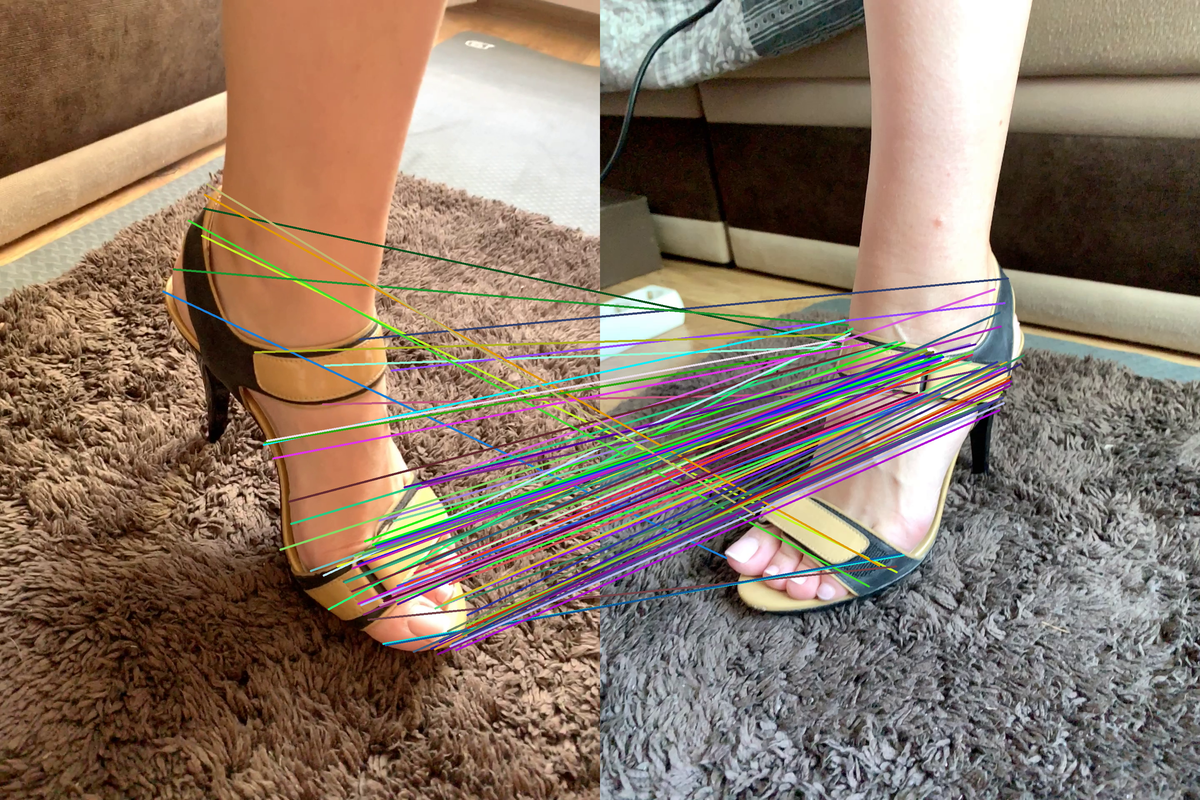} &
            \includegraphics[width=0.32\linewidth]{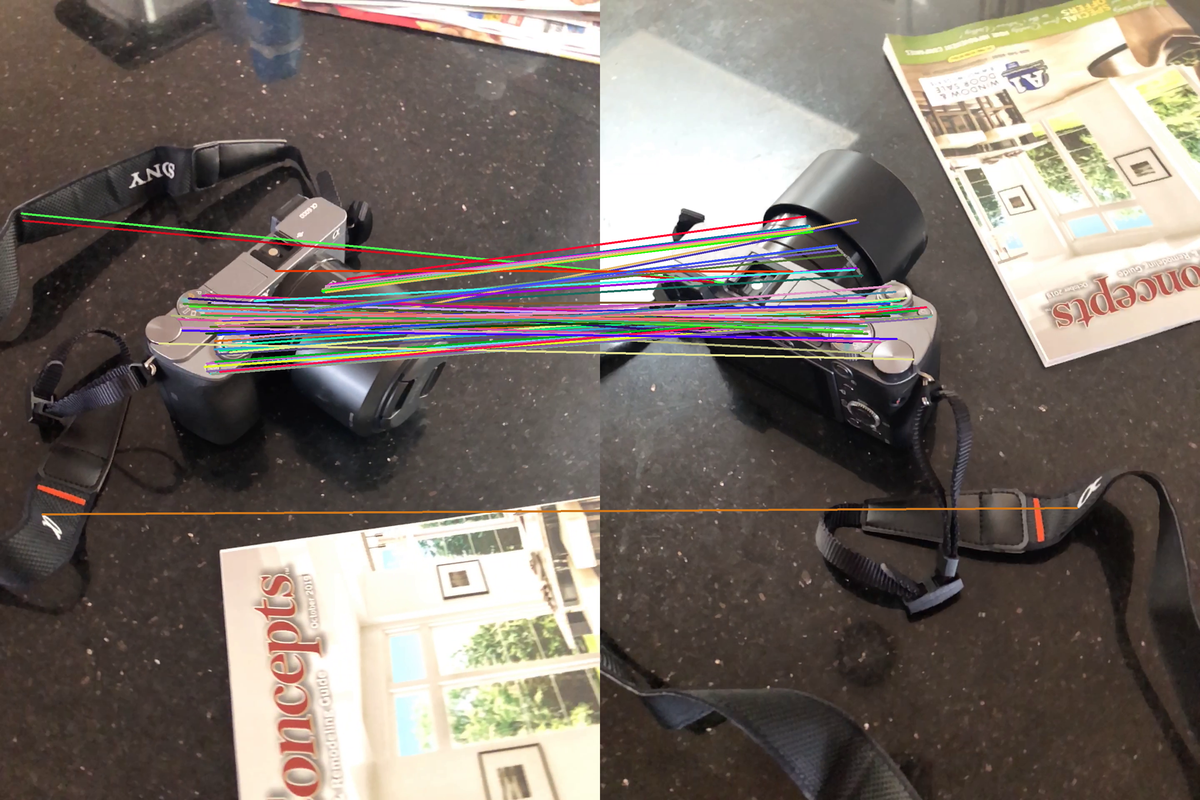}
        \end{tabular}
        \caption{Failure cases for LFM-3D.}
        \label{fig:qual_failure}
    \end{subfigure}
    
    \end{center}
    \vspace{-5pt}
    \caption{\textbf{Qualitative results for our LFM-3D method.} We show predicted correspondences with confidence threshold $0.2$. An absence of correspondence lines means that the model found no matches. \textbf{(a)} Correct matches are shown in green and incorrect matches ($>3$ pixel error) are shown in red. \textbf{(b)} \& \textbf{(c)} Ground truth isn't available, so we show matches in randomized colors.}
    \label{fig:qual}
\end{figure*}

\subsection{Results}

\noindent\textbf{Feature-level matching \& model ablations.}
Figure~\ref{fig:gso_pr_curves} presents precision/recall curves for LFM-3D on the Google Scanned Objects dataset for two object classes: \emph{shoe} (Fig.~\ref{fig:gso_pr_curves_a}) and \emph{camera} (Fig.~\ref{fig:gso_pr_curves_b}). 
We train class-specific matchers on only \textit{shoe} and \textit{camera} data and use NOCS and MDE as the 3D signals, respectively.
We compare LFM-3D against model variants that allow us to ablate the importance of the different components in our system: finetuning, usage of 3D signals, and improved positional encoding - all with a maximum of 1024 features.
To generate the curves, we sweep over confidence values in $[0, 1]$ to filter match predictions.
We use the following terminology for variants:

\begin{itemize}
    \item \textit{SG-MegaDepth} - Default SuperPoint + SuperGlue model.
    \item \textit{SG-Finetuned} - Default SuperPoint + SuperGlue model, initialized with \textit{SG-MegaDepth} and finetuned on Google Scanned Objects rendered pairs.
    \item \textit{SG + NOCS/MDE} - Finetuned SuperPoint + SuperGlue model, with an additional input encoding layer for 3D signals per keypoint (Eq.~\eqref{eq:keypoint-encoder-3d}). We use NOCS for \emph{shoe} and inverse depth estimates for \emph{camera}.
\end{itemize}

We observe that LFM-3D outperforms all baseline methods, with the most significant improvements being in maximum recall. 
The NOCS-based LFM-3D outperforms \textit{SG-Finetuned} on recall by $6.2\%$ and at maximum recall for \textit{SG-Finetuned} ($44.3\%$), it sees a $28.9\%$ increase in precision.
The significant gap between \textit{SG + MDE} and \textit{LFM-3D (MDE)} indicates that 
an effective positional encoding can enhance the model's capability to integrate low-dimensional 3D information.
Lastly, we note a large performance gap between the baseline \textit{SG-MegaDepth} method and all methods finetuned on synthetic renderings, indicating the importance of training learnable matchers on domain-specific data.

\noindent\textbf{Class-specific relative pose estimation.}
We evaluate relative pose estimation on the Objectron~\cite{objectron2021} dataset, by sampling pairs of video frames with camera axes angle distances between $90^{\circ}$-$120^{\circ}$. 
We limit keypoint extraction to a segmentation mask to minimize the domain gap.
We compare LFM-3D against conventional and state-of-the-art feature matching techniques; results are presented in Table~\ref{tab:rel_pose_results}.

Focusing on the first two columns of Table~\ref{tab:rel_pose_results}, we compare LFM-3D to all relevant benchmarks.
LFM-3D \textit{(class-specific)} and SuperPoint+SG-Finetuned \textit{(class-specific)} are both finetuned only on rendered pairs for their respective object classes.
As before, the \emph{shoe} LFM-3D model uses estimated NOCS maps and the \emph{camera} LFM-3D model uses estimated depth maps.
The LFM-3D model based on NOCS outperforms all other methods for the \emph{shoe} case.
Here, we observe a large gap between our method (row 11) and the finetuned 2D-only SuperPoint+SuperGlue (row 9): by $8.6$\% for Acc$@15^{\circ}$. 
More interestingly, we also outperform NOCS-based PnP variants by large margins (up to $10.4\%$ 
against the dense version (row 4)), which shows the limitations of directly using estimated NOCS maps for relative pose prediction.
Similarly, our depth-based \emph{camera} LFM-3D model outperforms all other class-specific methods. Although we observe a less drastic improvement over the 2D-only techniques, we note that our method can directly use an off-the-shelf MDE model to provide up to $1.5\%$ improvement.

\noindent\textbf{Relative pose estimation on novel classes.}
The \textit{generic} variants of SuperPoint+SG-Finetuned and LFM-3D (MDE) are trained on the entire GSO catalog of objects, including \textit{shoes} and \textit{cameras}. 
Results on the five other Objectron categories are aggregated into a single set of \textit{unseen} results, but per-class results are included in the appendix.
We note that the generic LFM-3D MDE model (row 13) outperforms the generic SuperPoint+SG-Finetuned (row 10) on \emph{all seen and unseen object classes}, by an average of $1.4\%$ for Acc@5$^{\circ}$ 
for unseen objects. Even further, generic LFM-3D outperforms (row 13) the class-specific LFM-3D camera model (row 12).
From these results, we observe that LFM-3D demonstrates an ability to improve from large amounts of data. The strong performance on the \textit{unseen} category shows that LFM-3D can also generalize well.

\noindent\textbf{Qualitative results and limitations}
Figure~\ref{fig:qual} shows how LFM-3D proposes improved correspondences between images with relatively small overlapping visible regions. 
\textit{SG-MegaDepth} suffers from low confidence predictions under very-wide baselines, but still correctly predicts many correspondences for minor viewpoint changes (Figure~\ref{fig:qual_gso}, bottom left). Finetuning SuperGlue results in higher-confidence predictions, but adding 3D signals with positional encoding leads to more correct matches and a wider distribution of proposed correspondences across the object's geometry (Figure~\ref{fig:qual_objectron}), which simplifies essential matrix estimation.

The disparity between LFM-3D performance on NOCS and MDE shows that the model can be sensitive to the choice of 3D signal. Figure~\ref{fig:qual_failure} shows examples where LFM-3D still struggles to match reliably. In particular, we observe that feature-less objects (first column) negatively affect local feature extractors and 3D signal estimators alike. Out-of-distribution geometries (heel shoes in the second column, camera strap in the third column) can also confuse the geometry estimation model and match poorly at inference time.
\section{Conclusions}
\label{sec:conclusions}
In this work, our main contribution is a novel method for local feature matching that integrates estimated 3D signals to guide correspondence estimation -- leading to high-quality feature associations under wide baseline conditions.
Experiments indicate that the way the 3D signal is incorporated into the model matters, with improvements observed when using positional encodings based on periodic functions.
Our enhanced feature correspondences enable a significant boost in performance for relative pose estimation.%

\appendix
\section*{Appendix}

\section{Comparison to additional baselines}

\begin{table*}
    \small
    \centering
    \begin{tabular}{lccc}
        \toprule
        & \multicolumn{3}{c}{\emph{Shoe}} \\
        Method & Acc@$5^\circ$ & Acc@$10^\circ$ & Acc@$15^\circ$ \\ %
        \midrule
        
        Official SP+SG release & 2.3 & 4.8 & 7.2 \\ %
        SuperPoint + SuperGlue (ours) & 2.6 & 5.6 & 8.6 \\ %
        \midrule
        
        SP+SG w/ NOCS filtering ($d = 0.03$) & 8.4 & 18.5 & 25.3 \\ %
        SP+SG w/ NOCS filtering ($d = 0.05$) & 9.6 & 19.7 & 27.6 \\ %
        SP+SG w/ NOCS filtering ($d = 0.1$) & 10.5 & 21.5 & 29.7 \\ %
        SP+SG w/ NOCS filtering ($d = 0.2$) & 10.8 & 22.1 & 30.6 \\ %
        SP+SG w/ NOCS filtering ($d = 0.5$) & 10.8 & 22.2 & 30.6 \\ %
        \midrule

        SP + LFM-3D (copied from Table 1) & \textbf{14.3} & \textbf{29.3} & \textbf{40.2} \\ %
        \bottomrule
    \end{tabular}
    
    \caption{\textbf{Additional baselines for relative pose estimation on the \emph{shoe} Objectron sampled video frame pairs.} We present results for relative pose estimation in terms of Accuracy@$N$. At the bottom, we copy the LFM-3D results from Table 1.}
    \label{tab:rel_pose_baselines_appendix}
\end{table*}

 In this section, we compare our method to some other baseline methods for relative pose estimation. Table~\ref{tab:rel_pose_baselines_appendix} reports the same relative pose metrics as Table 1 for four additional baselines:
 
 \begin{itemize}
     \item \textit{Official SP+SG release} - We use the official SuperGlue weights provided by the authors on \href{https://github.com/magicleap/SuperGluePretrainedNetwork}{github}. 
     \item \textit{SP+SG w/ NOCS filtering} - We use the SuperPoint+SuperGlue (finetuned) model to propose candidates, and only keep correspondences whose NOCS map values are within distance $d$ of one another.
 \end{itemize}

\noindent LFM-3D outperforms all of the additional baselines for the three challenging Acc@N error thresholds.

\paragraph{Comparison to Official SuperGlue baseline.}
Note that these results (\textit{Official SP+SG}) are roughly equivalent to our own SuperPoint+SuperGlue model trained on Megadepth, without finetuning (i.e. SuperPoint + SuperGlue (ours)). This indicates that we have reproduced SuperGlue in Tensorflow successfully.
 
\paragraph{Comparison to NOCS-based heuristics baseline.} Rather than integrating NOCS 3D signals in a learned manner, we experimented with using a heuristics-based approach for filtering correspondences proposed by the 2D-only method by NOCS distance (\textit{SP + SG w/ NOCS filtering}). This naive method of integrating NOCS coordinates underperforms LFM-3D, predictably. We note that LFM-3D's greatest improvements are in increased recall, as shown by Figure 4 in the main paper. This filtering method cannot introduce new correspondences to the limited set produced by the 2D-only baseline model.

\section{Unseen per-category pose estimation results.}
Table~\ref{tab:rel_pose_unseen_results} presents per-class results on the 5 \textit{unseen} object classes considered during our pose estimation experiments. We notice significant variation in pose estimation results across all methods, but consistently observe that LFM-3D (generic, MDE) outperforms the generic finetuned 2D baseline.
\begin{table*}
    \small
    \centering
    \resizebox{\textwidth}{!}{
    \begin{tabular}{lccc|ccc|ccc|ccc|ccc}
        \toprule
        & \multicolumn{3}{c}{\emph{Bike}} & \multicolumn{3}{c}{\emph{Book}} & \multicolumn{3}{c}{\emph{Cereal box}} & \multicolumn{3}{c}{\emph{Chair}} & \multicolumn{3}{c}{\emph{Laptop}} \\
        Method & Acc@$5^{\circ}$ & Acc@$10^{\circ}$ & Acc@$15^{\circ}$ & Acc@$5^{\circ}$ & Acc@$10^{\circ}$ & Acc@$15^{\circ}$ & Acc@$5^{\circ}$ & Acc@$10^{\circ}$ & Acc@$15^{\circ}$ & Acc@$5^{\circ}$ & Acc@$10^{\circ}$ & Acc@$15^{\circ}$ & Acc@$5^{\circ}$ & Acc@$10^{\circ}$ & Acc@$15^{\circ}$\\
        \midrule
        
        SIFT + MNN & 3.7 & 7.6 & 11.1 & 8.8 & 15.1 & 19.6 & 4.2 & 7.9 & 11.0 & 2.7 & 5.5 & 8.1 & 4.6 & 9.7 & 14.7 \\
        SIFT + NN/Ratio & 4.1 & 7.9 & 11.8 & 6.0 & 11.2 & 15.6 & 3.9 & 7.4 & 10.8 & 3.2 & 6.4 & 9.5 & 4.7 & 9.7 & 14.6 \\
        \midrule
        
        NOCS + Sparse-PnP & - & - & - & - & - & - & - & - & - & - & - & - & - & - & - \\
        NOCS + Dense-PnP & - & - & - & - & - & - & - & - & - & - & - & - & - & - & - \\
        \midrule
        
        LoFTR & 0.6 & 1.4 & 2.1 & 6.3 & 7.9 & 9.3 & 10.7 & 14.1 & 16.1 & 2.8 & 4.8 & 6.2 & 1.7 & 3.4 & 4.7 \\
        \midrule
        
        SuperPoint + MNN & 4.4 & 8.6 & 13.0 & 10.5 & 17.1 & 21.9 & 7.7 & 13.1 & 17.1 & 3.1 & 6.3 & 9.3 & 3.6 & 8.3 & 12.9 \\
        SuperPoint + NN/Ratio & \textbf{4.7} & \textbf{9.3} & \textbf{13.8} & 8.1 & 14.3 & 19.2 & 5.0 & 9.6 & 13.7 & 3.3 & 6.7 & 10.2 & 3.5 & 7.8 & 12.4 \\
        SuperPoint + SG-MegaDepth & 2.1 & 4.4 & 6.6 & 10.9 & 14.6 & 17.2 & 13.0 & 17.8 & 20.6 & 2.5 & 5.2 & 7.6 & 3.4 & 7.0 & 10.5 \\
        SuperPoint + SG-Finetuned \textit{(class specific)} & - & - & - & - & - & - & - & - & - & - & - & - & - & - & - \\
        SuperPoint + SG-Finetuned \textit{(generic)} & 3.8 & 7.7 & 11.6 & 33.4 & 48.2 & 53.8 & 35.5 & 46.5 & 51.6 & 5.0 & 9.6 & 13.7 & \textbf{10.6} & 22.6 & 30.9 \\
        \midrule
        \midrule
        
        SuperPoint + LFM-3D \textit{(class specific, NOCS)} & - & - & - & - & - & - & - & - & - & - & - & - & - & - & - \\
        SuperPoint + LFM-3D \textit{(class specific, MDE)} & - & - & - & - & - & - & - & - & - & - & - & - & - & - & - \\
        \midrule
        SuperPoint + LFM-3D \textit{(generic, MDE)}  & 3.9 & 7.9 & 12.0 & \textbf{35.5} & \textbf{50.7} & \textbf{56.2} & \textbf{40.0} & \textbf{52.0} & \textbf{57.2} & \textbf{5.6} & \textbf{10.6} & \textbf{14.9} & \textbf{10.6} & \textbf{23.6} & \textbf{32.2} \\
        \bottomrule
    \end{tabular}
    }
    
    \caption{\textbf{Per-category welative pose estimation results} on the 5 \textit{unseen} Objectron object classes.}
    \label{tab:rel_pose_unseen_results}
\end{table*}

\section{Ablation on number of keypoints}
\begin{table}[]
    \small
    \centering
    \begin{tabular}{lccc}
        \toprule
        & \multicolumn{3}{c}{\emph{Shoe}} \\
        Method & Acc@$5^\circ$ & Acc@$10^\circ$ & Acc@$15^\circ$ \\
        \midrule
        SP (1024) + SG & 11.0 & 22.2 & 31.6 \\
        SP (2048) + SG & 11.4 & 22.6 & 31.3 \\
        SP (4096) + SG & 10.7 & 21.6 & 30.6 \\
        \midrule
        SP (1024) + LFM-3D & 14.3 & 29.3 & 40.2 \\
        SP (2048) + LFM-3D & \textbf{15.5} & \textbf{30.6} & \textbf{42.3} \\
        SP (4096) + LFM-3D & 13.9 & 28.7 & 39.6 \\
        \bottomrule
    \end{tabular}
    \caption{\textbf{Ablation study over number of extracted keypoints.} We compare the finetuned 2D-only baseline (SuperPoint~\cite{detone18superpoint} + SuperGlue~\cite{sarlin2020superglue}) to our LFM-3D model from NOCS estimates on the \emph{shoe} Objectron evaluation set.}
    \label{tab:num_keypoints_ablation}
\end{table}

Table~\ref{tab:num_keypoints_ablation} compares LFM-3D with the finetuned 2D-only baseline with varying numbers of keypoints (2048 and 4096, vs baseline 1024 from the original paper) during relative pose estimation on Objectron \emph{shoe} image pairs. Note that we do not retrain models for each new keypoint input size. Our results show that LFM-3D benefits from additional keypoints more than the 2D-only baselines for 2048 keypoints, but performance drops with 4096 keypoints. We hypothesize that our method can produce better features due to augmented 3D inputs, while in the 2D-only case, having more features can introduce noise and false matches.

\begin{figure*}[t]
    \begin{center}
    \vspace{7pt}
    \begin{tabular}{cc|cc|cc}
    \includegraphics[width=0.14\linewidth]{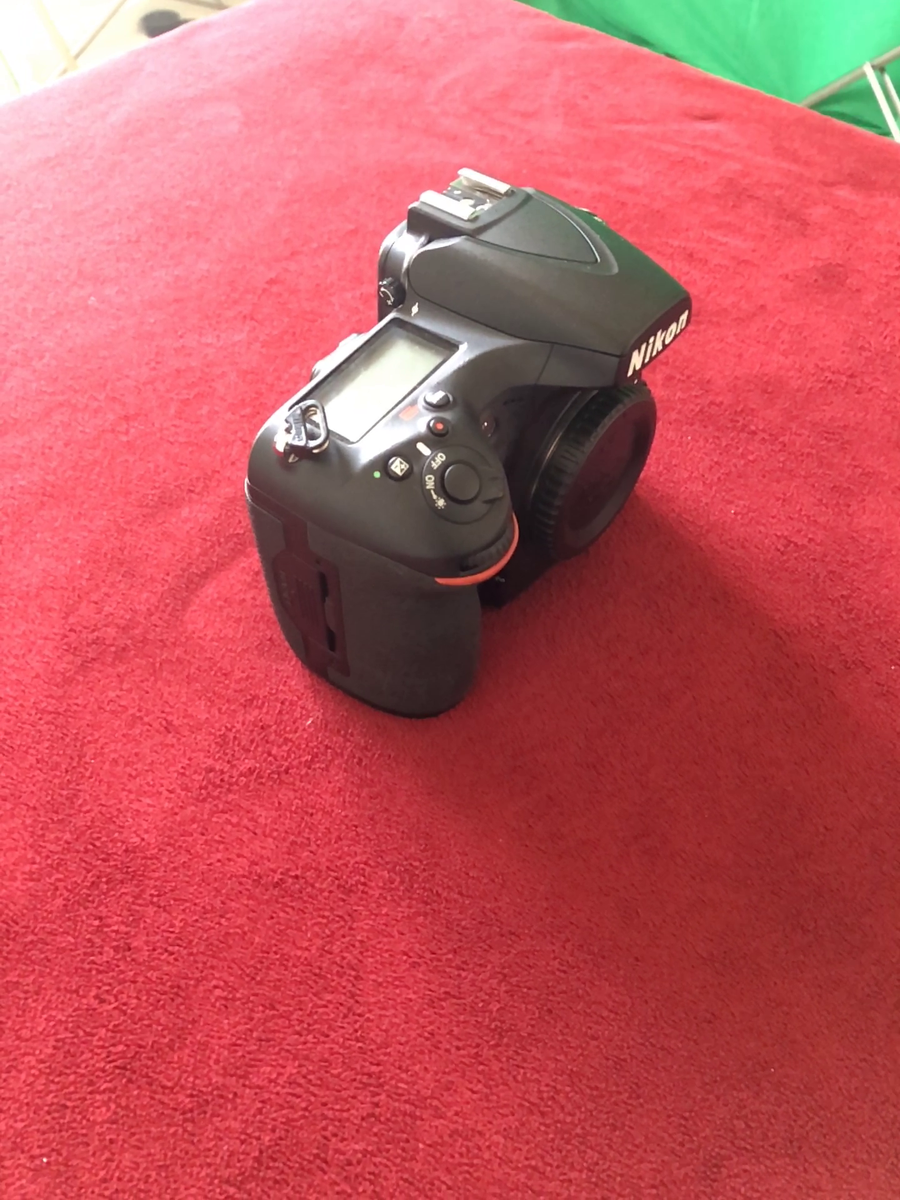} & 
    \includegraphics[width=0.14\linewidth]{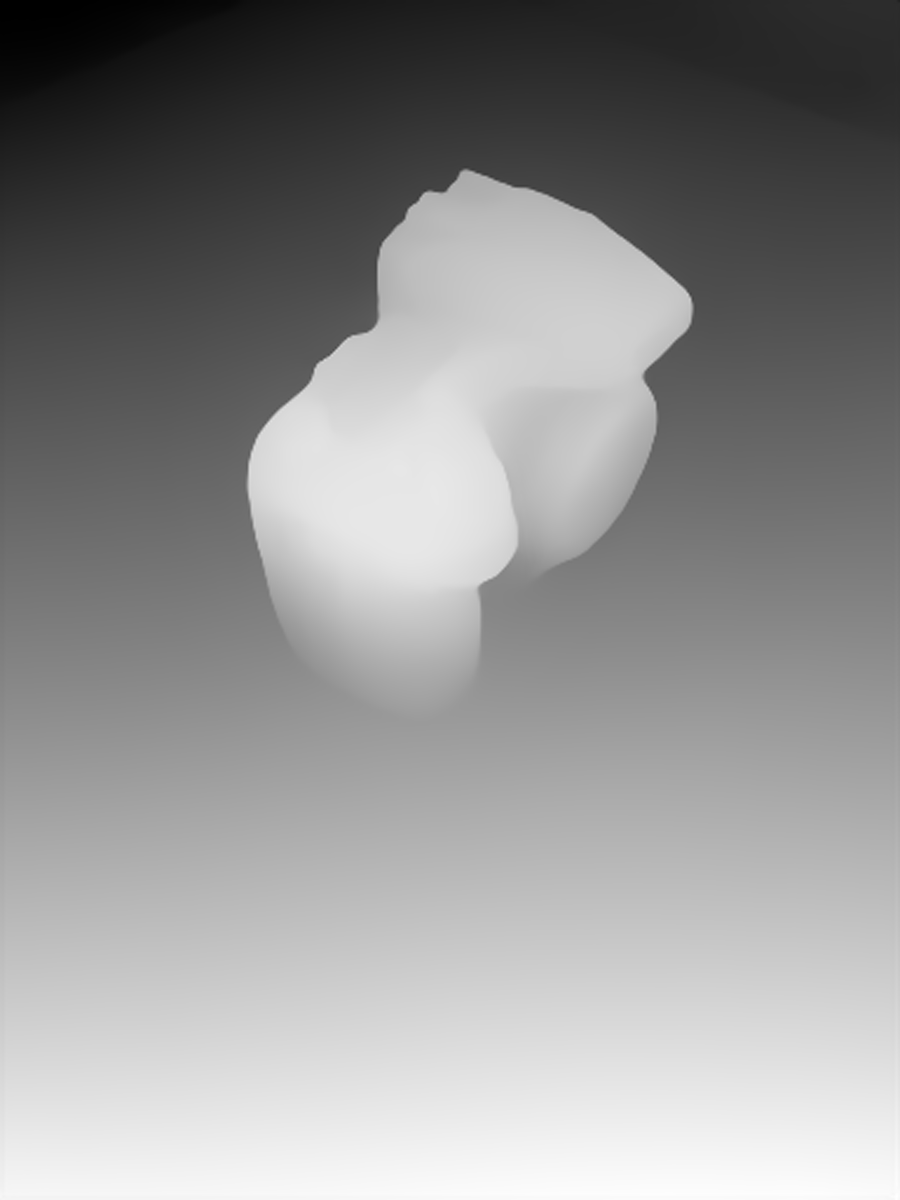} &
    \includegraphics[width=0.14\linewidth]{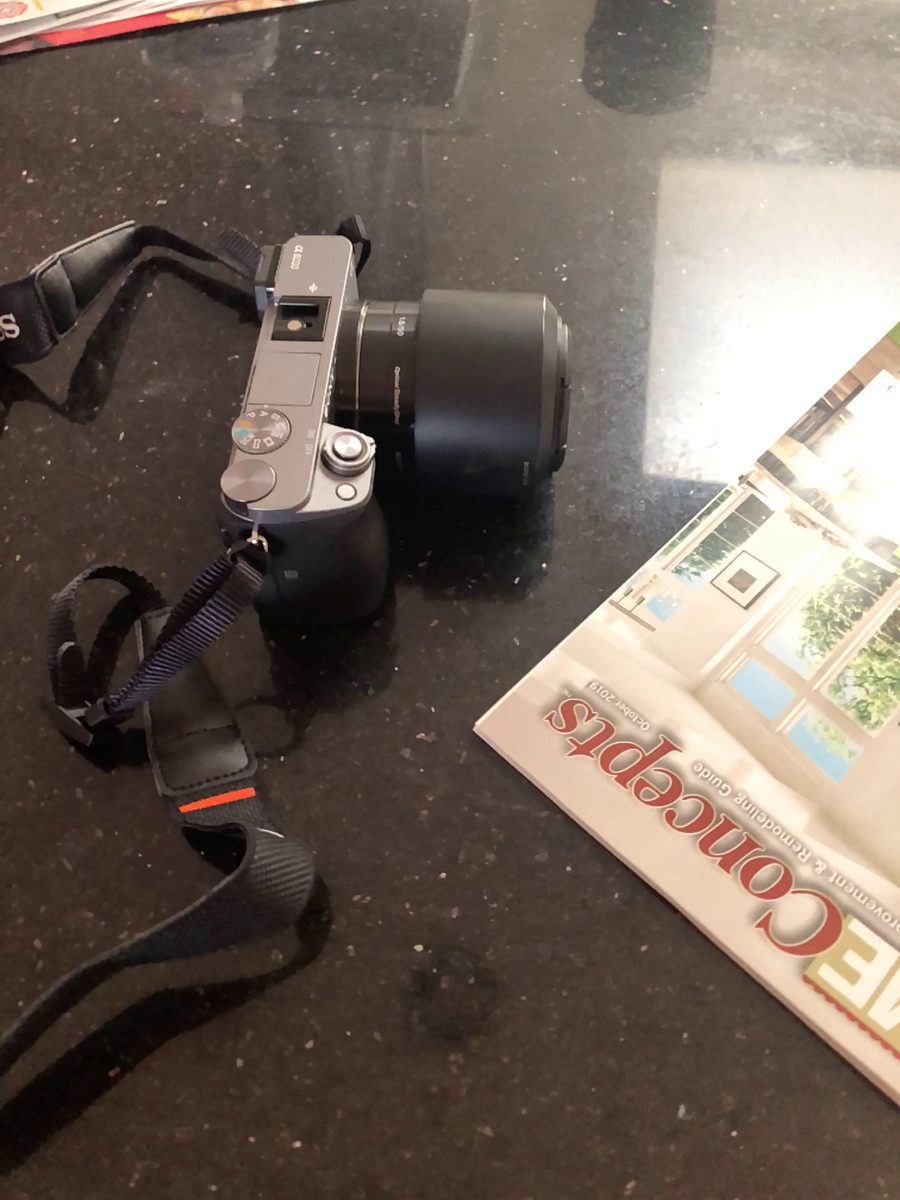} & 
    \includegraphics[width=0.14\linewidth]{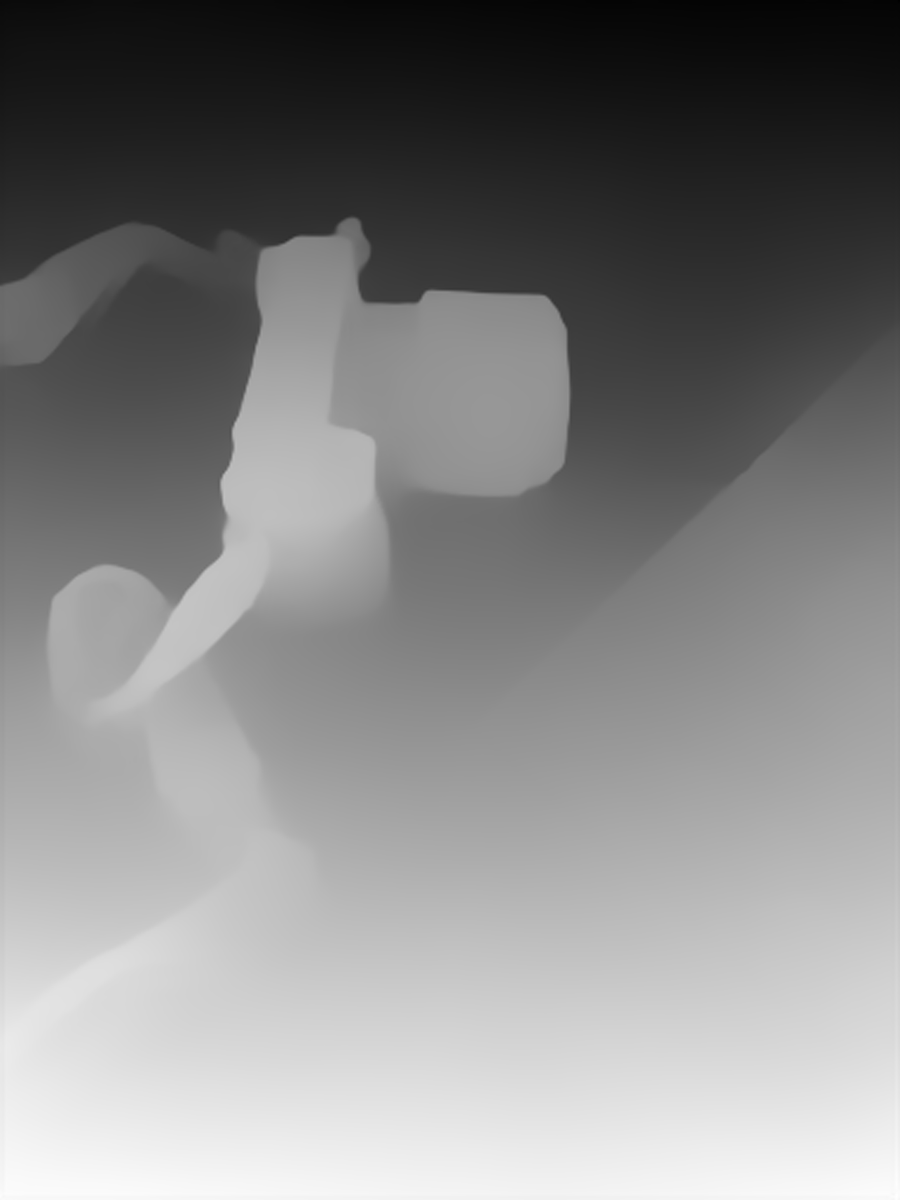} &
    \includegraphics[width=0.14\linewidth]{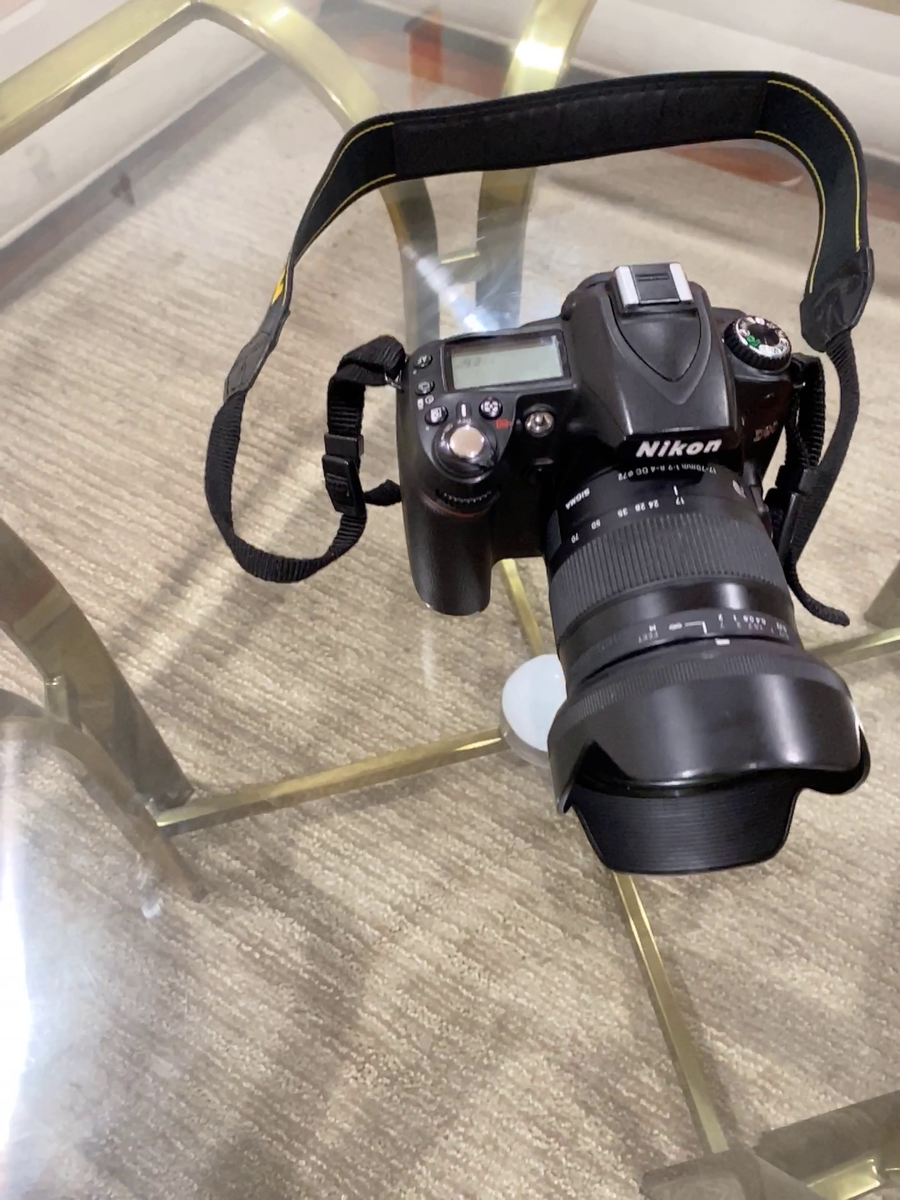} & 
    \includegraphics[width=0.14\linewidth]{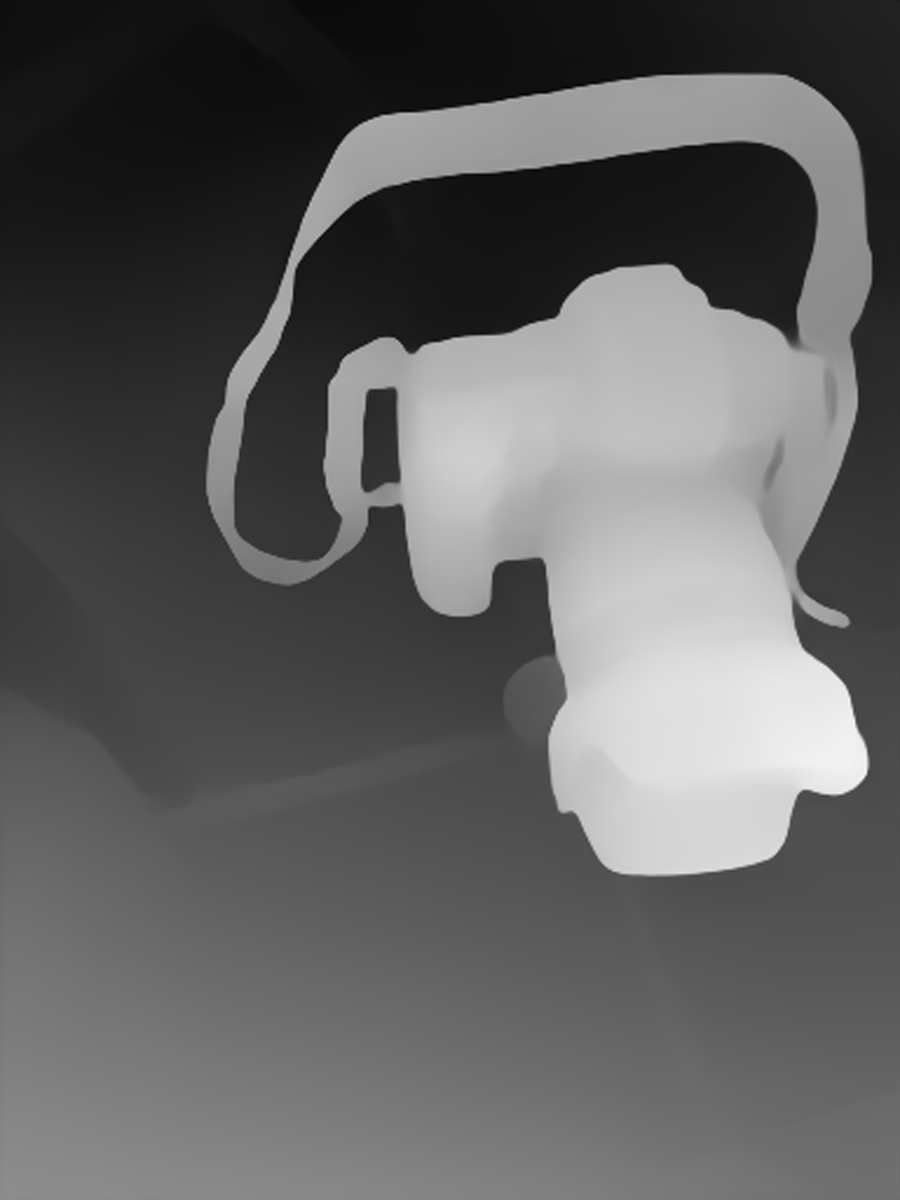}
    \end{tabular}
    \end{center}
    \vspace{-5pt}
    \caption{\textbf{Visualizations of monocular depth estimates on camera Objectron images.} Inverse depth values are generated using the general-purpose DPT~\cite{ranftl2021vision} monocular depth estimation model.}
    \label{fig:mde_examples}
\end{figure*}

\section{Examples of monocular depth estimates}
Figure~\ref{fig:mde_examples} presents qualitative examples from the general-purpose, pre-trained monocular depth estimation model, DPT~\cite{ranftl2021vision}. We use the published model weights available on their \href{https://github.com/isl-org/DPT}{github}.

\section{Additional qualitative examples}
\begin{figure*}[t]
    \begin{center}
    \hspace{-5pt}
    \includegraphics[width=\linewidth]{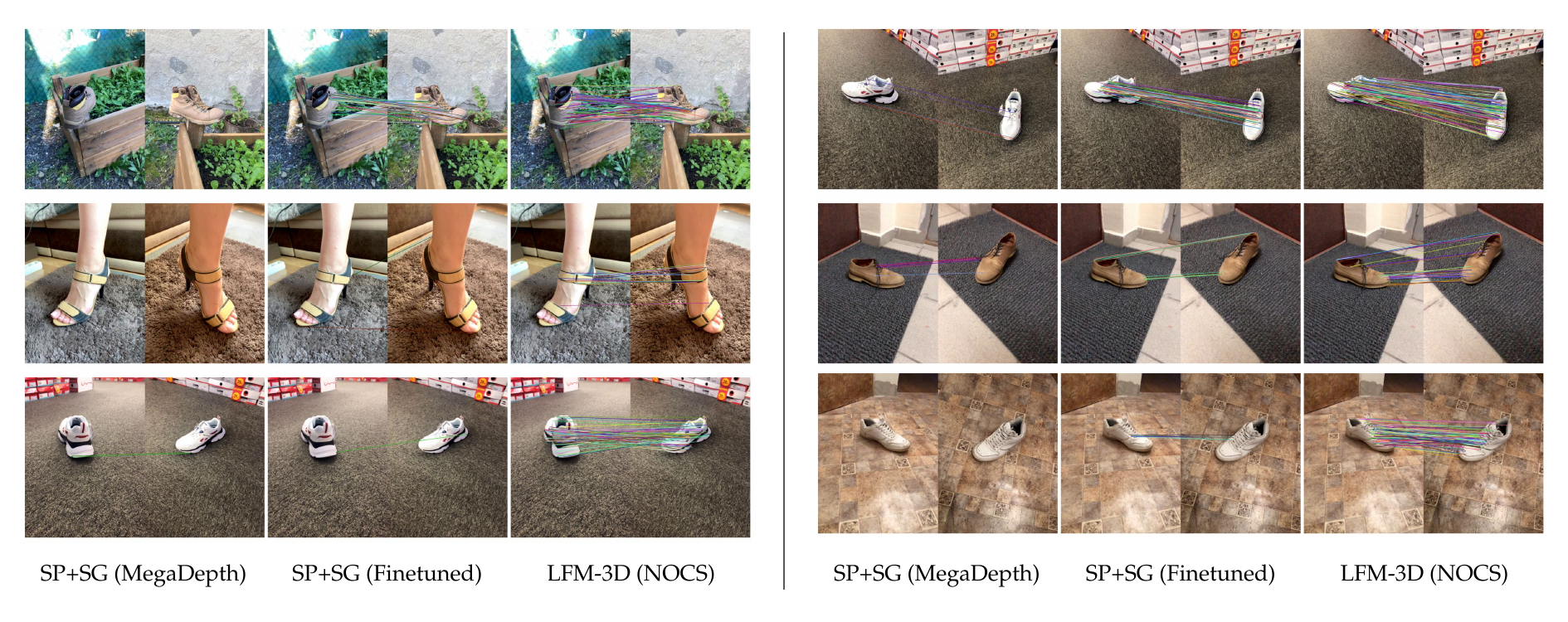}
    
    \end{center}
    \vspace{-5pt}
    \caption{\textbf{Additional qualitative examples from sampled Objectron frames.} For each triplet, we compare SP+SG-MegaDepth, SP+SG-Finetuned, and LFM-3D.}
    \label{fig:qual_appendix}
\end{figure*}

We present additional qualitative results on the Objectron dataset in Figure~\ref{fig:qual_appendix}. These results highlight that LFM-3D performs well across a variety of shapes, textures, and lighting conditions.

{
    \small
    \bibliographystyle{ieeenat_fullname}
    \bibliography{main}
}
\end{document}